\documentclass[10pt, a4paper]{article}
\usepackage{natbib}
\usepackage{booktabs} 
\usepackage{titlesec}
\usepackage[many]{tcolorbox}
\tcbuselibrary{skins,breakable,raster}
\usepackage{ragged2e} 
\usepackage{subcaption}
\usepackage{graphicx}
\usepackage{adjustbox} 
\newtcbox{\malehl}{on line, boxrule=0pt, arc=2pt,
  colback=blue!12, colframe=blue!12, left=2pt, right=2pt, top=0.2ex, bottom=0.2ex}
\newtcbox{\femalehl}{on line, boxrule=0pt, arc=2pt,
  colback=red!12,  colframe=red!12,  left=2pt, right=2pt, top=0.2ex, bottom=0.2ex}
\newtcbox{\neutralhl}{on line, boxrule=0pt, arc=2pt,
  colback=green!12, colframe=green!12, left=2pt, right=2pt, top=0.2ex, bottom=0.2ex}

\usepackage[final]{lrec2026} 

\titlespacing*{\paragraph}{0pt}{0.5ex plus .2ex minus .2ex}{1em}

\title{Gender Disambiguation in Machine Translation:\\Diagnostic Evaluation in Decoder-Only Architectures}

\name{%
\parbox{\textwidth}{\centering
Chiara Manna \quad Hosein Mohebbi \quad Afra Alishahi\\
Frédéric Blain \quad Eva Vanmassenhove}\vspace{0.5em}
}
\address{Research Center for Cognitive Science and Artificial Intelligence\\
         Tilburg School of Humanities and Digital Sciences\\
         Tilburg University, Netherlands\\
         \{c.manna, h.mohebbi, a.alishahi, f.l.g.blain, e.o.j.vanmassenhove\}@tilburguniversity.edu\\}
         
\abstract{
While Large Language Models achieve state-of-the-art results across a wide range of NLP tasks, they remain prone to systematic biases.
Among these, gender bias is particularly salient in MT, due to systematic differences across languages in whether and how gender is marked.
As a result, translation often requires disambiguating implicit source signals into explicit gender-marked forms. 
In this context, standard benchmarks may capture broad disparities but fail to reflect the full complexity of gender bias in modern MT. 
In this paper, we extend recent frameworks on bias evaluation 
by: (i) introducing a novel measure coined "Prior Bias", capturing a model's default gender assumptions, and (ii) applying the framework to decoder-only MT models.
Our results show that, despite their scale and state-of-the-art status, decoder-only models do not generally outperform encoder-decoder architectures on gender-specific metrics; however, post-training (e.g., instruction tuning) not only improves contextual awareness but also reduces the masculine Prior Bias. The code and supplementary data used in this work are available at \href{https://github.com/chiaramanna/gender-bias-diagnostics-mt}{github.com/chiaramanna/gender-bias-diagnostics-mt}.
 \\ \newline  }

\begin{document}

\maketitleabstract

\section{Introduction}
Recent years have seen remarkable advancements in the field of Machine Translation (MT), marked by the shift from statistical to Neural MT, and the introduction of Transformers \citep{attention2017}, which have since become the backbone of most NLP systems.
While early implementations relied on an encoder-decoder design \citep{tiedemann-thottingal-2020-opus, liu-etal-2020-multilingual-denoising,nllb200, raffel2020t5}, current state-of-the-art systems are predominantly decoder-only Large Language Models (LLMs) \citep{openai2023gpt4, touvron2023b, gemmateam2024gemma,jiang2024mixtral}, pretrained on massive text corpora and often fine-tuned for specific applications \citep{bommasani2021opportunities}. Their success largely stems from the ability to build contextualized representations and capture long-range dependencies \citep{attention2017, tay2020long}, which has led to substantial improvements in translation quality and established LLMs as the new state-of-the-art in MT \citep{kocmi-etal-2023-findings, kocmi-etal-2024-findings, deutsch2025wmt24pp}. At the same time, awareness of their limitations has grown in parallel \citep{kirk2021bias, bender2021parrots, navigli2023biases, gallegos2024bias}. 
\newtcolorbox{LangBox}[1]{title=\textsc{#1}, width=\columnwidth,
  colback=gray!8, colframe=gray!8, arc=6pt, boxrule=0.6pt, colbacktitle=gray!30, 
  left=2pt, right=2pt, top=4pt, bottom=4pt,
  fonttitle=\bfseries\footnotesize, fontupper=\small\centering,
  before skip=2pt, after skip=2pt}
  
\newtcolorbox{PromptBox}[1][]{
  enhanced, breakable,
  colback=gray!8, colframe=gray!8,
  colbacktitle=gray!30,
  boxrule=0.6pt, arc=6pt,
  left=2pt, right=2pt, top=6pt, bottom=6pt,
  fontupper=\ttfamily\small,
  title=#1
}

\begin{figure}[h!]
\centering
\begin{LangBox}{Input}
The cook prepared a soup for the \textbf{housekeeper} because \malehl{\textbf{he}} helped clean the room.
\end{LangBox}
\begin{LangBox}{Output}
Il cuoco ha preparato una zuppa per \femalehl{\textbf{la governante}} perché ha aiutato a pulire la stanza.
\end{LangBox}
\vspace{0.25em}

\begin{LangBox}{Input}
The \textbf{farmer} offered apples to the housekeeper because \femalehl{\textbf{she}} had too many of them.
\end{LangBox}
\begin{LangBox}{Output}
\malehl{\textbf{Il contadino}} offrì mele alla governante perché ne aveva troppe.
\end{LangBox}
\vspace{0.5em}
\caption[]{\small{EN$\rightarrow$IT WinoMT examples showing outputs\footnotemark{} defaulting to \textit{la governante} (\femalehl{fem.}) and \textit{il contadino} (\malehl{masc.}), despite the gender cue (\textit{he}/\textit{she}).
}}\label{fig_1:gender_bias_example}
\end{figure}
\footnotetext{Generated by ChatGPT on September 29th, 2025.}
\newpage
Their growing scale and complexity make it increasingly difficult to trace back decisions or explain systematic errors. Moreover, because they are pretrained on largely uncurated corpora, with little transparency about data composition, it is nearly impossible to ensure high-quality training data. This lack of control raises concerns about the biases that such models inherit, propagate, and potentially amplify, as they become increasingly integrated into one’s everyday life \citep{bansal2022survey}. 

Despite extensive research efforts on fairness and bias in NLP \citep{sun-etal-2019-mitigating, costa2019analysis,blodgett-etal-2020-language, stanczak2021survey}, one form of bias that remains particularly salient in MT is gender bias, due to systematic differences across languages in whether and how gender is marked~\citep{ackerman, cao-daume}: while English primarily marks gender through pronouns, morphologically richer languages such as Italian or German require gender agreement across multiple parts of speech \citep{stahlberg}. This often forces implicit source information to become explicit in the target \citep{vanmassenhove-etal-2018-getting, vanmassenhove2024gender}. In principle, contextual cues such as gendered pronouns could guide models' decisions, yet in practice, these are often disregarded in favor of stereotypical associations. Figure~\ref{fig_1:gender_bias_example} illustrates this through two examples from the WinoMT dataset \citep{winomt}, where the corresponding translations overlook the contextual cues available: \textit{housekeeper} is realized in the feminine form despite the masculine pronoun \textit{he}, and \textit{farmer} in a masculine one despite the feminine pronoun \textit{she}. Such errors not only reproduce gender inequalities but also risk reinforcing them on a broader scale \citep{blodgett-etal-2020-language}.

Standard gender accuracy metrics reveal broad disparities in bias evaluation but offer little insight into whether models actually use available contextual gender cues or simply default to stereotypes. 
To address this, we previously propose an evaluation framework based on Minimal Pair Accuracy (MPA) \citep{manna-etal-2025-paying}, which measures how consistently models integrate contextual gender cues when translating. Specifically, we consider a translation as "accurate" only if swapping the gender cue (e.g. he/she) yields correct translations in both cases, that is, within a minimal pair. 

Building on this idea, we extend the framework along two dimensions.
First, we introduce \textbf{Prior Bias}, a novel measure to quantify a model’s gender assumption in the absence of explicit cues. To this end, we constructed a Neutral Set, the neutralized extension of WinoMT \citep{winomt} where gendered pronouns are replaced with \textit{they/them} and agreement is adjusted accordingly (e.g. adjectives/verbs). 
Combined with MPA~\citep{manna-etal-2025-paying}, this extension allows us to diagnose whether model errors stem from ignoring contexts, defaulting to a prior bias, or both. Secondly, we further apply the extended framework to \textbf{decoder-only models}, evaluating and comparing gender bias in MT across both foundational and instruction-tuned variants. We find that decoder-only models do not surpass encoder-decoder ones on our gender-specific MT metrics; however, post-training (e.g. instruction tuning) improves contextual cue integration and reduces Prior Bias.


\section{Related Work}

Over the past decade, research on gender bias in NLP has received increasing attention. In 2014, \citet{Schiebinger2014} exposed the risks of overlooking gender in science and technology. As an early example, they pointed to Machine Translation (MT), noting that automatic translation tools frequently defaulted to masculine pronouns, reflecting their disproportionate frequency in training data. \\ 
Since then, systematic gender bias has been documented across a wide range of domains, from literature \citep{hoyle-etal-2019-unsupervised}, news \citep{wevers-2019-using}, and media \citep{asr2021gendergap}, to word embeddings \citep{bolukbasi2016man}, pretrained language models \citep{nadeem-etal-2021-stereoset}, and downstream tasks, such as sentiment analysis \citep{kiritchenko-mohammad-2018-examining}, coreference resolution \citep{rudinger-etal-2018-gender, webster-etal-2018-mind,zhao-etal-2018-gender}, hate speech detection \citep{park-etal-2018-reducing}, language generation \citep{sheng-etal-2020-towards}, syntactic analysis \citep{garimella-etal-2019-womens} and, indeed, MT \citep{vanmassenhove-etal-2018-getting, savoldi2021gender}. 
The latter remains a particularly challenging setting due to the cross-linguistic nature of the task \citep{stahlberg,ackerman, cao-daume}. Translating from languages where gender is sparsely marked (e.g., English) into morphologically richer ones (e.g., German, Italian or Hindi) often requires making implicit information explicit in the target \citep{vanmassenhove-etal-2018-getting, vanmassenhove2024gender}. This added complexity increases the likelihood that models may rely on (stereotypical) statistical associations when resolving gender, as illustrated in Figure~\ref{fig_1:gender_bias_example}. 

Research on gender bias in MT has primarily focused on analyzing system outputs \citep{rescigno2020case,ramesh2021evaluating} and developing bias mitigation strategies, targeting different stages of the translation pipeline. These include rewriting translations into explicitly gendered forms \citep{vanmassenhove-etal-2018-getting,moryossef2019filling,habash2019automatic} or neutral alternatives \citep{vanmassenhove-etal-2021-neutral,sun2021they}, applying embedding-level debiasing methods \citep{hirasawa2019debiasing,font2019equalizing}, and leveraging counterfactual data augmentation \citep{zmigrod2019counterfactual} or domain adaptation \citep{saunders2020reducing}. Nonetheless, these approaches have generally proven only partially effective and difficult to scale to modern MT systems \citep{savoldi2025decade}. The challenge has been further amplified by the rise of LLMs, which now underpin the state-of-the-art in MT \citep{kocmi-etal-2023-findings, kocmi-etal-2024-findings, deutsch2025wmt24pp}, alongside most NLP tasks. Built on the Transformer architecture \citep{attention2017}, these models excel in constructing contextualized token representations by capturing long-range dependencies \citep{tay2020long}, but their opacity raises concerns about inherited and amplified biases \citep{caliskan2017semantics, kirk2021bias, bender2021parrots, navigli2023biases, gallegos2024bias}. Moreover, evidence shows that Transformer-based models for Machine Translation often fail to exploit context effectively. Studies on intra- and inter-sentential context usage \citep{goindani-shrivastava-2021-dynamic,voita-etal-2021-analyzing,sarti2024_context,mohammed-niculae-2024-measuring} reveal that models frequently ignore relevant context and/or attend to irrelevant tokens, resulting in systematic errors and biased outputs \citep{kim-etal-2019-document,yin-etal-2021-context}.

Within this landscape, available benchmark datasets, such as the WinoMT \citep{winomt,bentivogli-etal-2020-gender}, have provided controlled settings to expose systematic gaps across systems and language pairs, but are insufficient to capture the full complexity of gender bias in modern, state-of-the-art MT. Standard evaluation frameworks largely operate at the surface level, verifying whether gender markings in the output match the source (Figure~\ref{fig_1:gender_bias_example}). However, no insights into the mechanisms underlying model decisions are revealed. Particularly, \emph{we cannot determine whether models rely on contextual cues or default to stereotypes, nor can we capture prior gender assumptions}. These limitations underscore the need for more fine-grained evaluation frameworks that combine evaluative and diagnostic purposes, by quantifying disparities in performance while revealing where errors arise from.
\section{Experimental Setup}
We investigate whether decoder-only LLMs that underpin the current state-of-the-art in MT integrate contextual gender cues during translation and to what extent their outputs reflect a default (prior) gender preference. To do so, we maintain the two dimensions proposed in the original framework \citep{manna-etal-2025-paying}: an output-level evaluation, which examines model behavior through translation outputs; and an attention-based analysis, which probes internal attention patterns as a proxy for cue integration. This section describes the models, prompting strategies, and data used in our experiments, while the detailed setup for each dimension is presented in its respective section.

\subsection{Models}
We select three decoder-only models of identical size and architecture (7B parameters, 32 layers, 32 attention heads). 
\begin{enumerate}
    \item \textbf{Llama2}\footnote{\href{https://huggingface.co/meta-llama/Llama-2-7b-hf}{huggingface.co/meta-llama/Llama-2-7b-hf}} \citep{touvron2023b}: a general-purpose pretrained LLM used here as a baseline to assess gender bias in the absence of MT-specific adaptation.
    \item \textbf{TowerBase}\footnote{\href{https://huggingface.co/Unbabel/TowerBase-7B-v0.1}{huggingface.co/Unbabel/TowerBase-7B-v0.1}}: a variant of (1) that underwent \textbf{continued pretraining} \citep{alves2024tower} on a mixture of monolingual and parallel corpora, following standard MT training practices.
    \item \textbf{TowerInstruct}\footnote{\href{https://huggingface.co/Unbabel/TowerInstruct-7B-v0.2}{huggingface.co/Unbabel/TowerInstruct-7B-v0.2}}: an extension of (2) refined through \textbf{instruction tuning} on translation-specific tasks, achieving GPT-4-level performance on several MT benchmarks \citep{alves2024tower}. 
\end{enumerate}
We opt for \textbf{TowerLLM}, one of the few open-weight decoder-only model suites explicitly tuned for MT, with its instruction-tuned variant (3) consistently outperforming other open alternatives across MT benchmarks and competing with closed-source systems like GPT-4~\citep{openai2023gpt4}, even at the scale of 7B parameters~\citep{alves2024tower}. Moreover, the TowerLLM model suite enables a direct comparison between a general-purpose LLM, our base model (1), and its translation-tuned variants (2--3) under \textbf{controlled post-training conditions}, while retaining a realistic model size for practical MT deployment. \\

For context, we compare our results to the previously studied encoder-decoder systems, as reported in \citet{manna-etal-2025-paying}.

\begin{figure}[h!]
\centering
\begin{PromptBox}[Llama2 / TowerBase]
Translate the following text from English into Italian.\\
English: \{source\_sentence\}\\
Italian: \{generated\_translation\}
\end{PromptBox}

\begin{PromptBox}[TowerInstruct]
<|im\_start|> user\\
Translate the following text from English into Italian.\\
English: \{source\_sentence\}\\
Italian: <|im\_end|>\\
<|im\_start|> assistant\\
\{generated\_translation\}
\end{PromptBox}
\caption{\small
Prompt templates for the EN$\rightarrow$IT task. Plain for Llama2 and TowerBase, chat-style for TowerInstruct.}
\label{fig:prompt_templates}
\end{figure}

\subsection{Prompting}\label{sec:prompt}
MT-specific prompting strategies vary widely across models and language pairs, and no single template has yet proven consistently optimal \citep{zhang2023promptingmt, leiter2024prexme}. While prior work has predominantly focused on few-shot prompting \citep{hendy2023gptmt, vilar2023prompting}, our goal is to isolate model behavior rather than to optimize performance through prompt engineering. We therefore adopt the (zero-shot) prompt format used during TowerInstruct’s instruction tuning \citep{alves2024tower}, and provide a plain natural language equivalent for Llama2 and TowerBase, which were not trained with predefined prompt templates. The prompt templates used ensure a consistent and comparable setup across the evaluated models (see Fig.~\ref{fig:prompt_templates}).

\subsection{Data}
We rely on the WinoMT challenge set \citep{winomt}, which builds on Winogender \citep{rudinger-etal-2018-gender} and WinoBias \citep{zhao-etal-2018-gender} as a benchmark for gender bias in MT. Each sentence follows a simple template, containing (i) an ungendered profession noun (the target noun to be translated) and (ii) a gendered pronoun that provides the gold gender cue (see Figure~\ref{fig_1:gender_bias_example}). This includes a:  
\begin{itemize}
    \item \textbf{Regular set:} 3,888 sentences balanced across masculine and feminine referents.
    \item \textbf{Pro-stereotypical and anti-stereotypical subsets:} 1,584 instances each, pairing professions with gendered pronouns aligning with or contradicting gender-role expectations (e.g., \textit{the housekeeper ... she} vs.\ \textit{the housekeeper ... he}).
    \end{itemize}
    \vspace{-0.3cm}
\begin{figure}[h!]
\centering
\begin{LangBox}{Gendered}
The \textbf{developer} visited the hairdresser because \malehl{\textbf{he}} needed to cut \malehl{\textbf{his}} hair.
\end{LangBox}
\begin{LangBox}{Neutralized}
The \textbf{developer} visited the hairdresser because \neutralhl{\textbf{they}} needed to cut \neutralhl{\textbf{their}} hair.
\end{LangBox}
\vspace{0.75em}
\caption{\small Example illustrating the neutralization process followed to construct the Neutral Set.}
\label{fig:neutralized_example}
\end{figure}
\vspace{0.3cm}
\noindent \textbf{Neutral set:} Following the approach of \citet{vanmassenhove-etal-2021-neutral}, we additionally release a neutralized version of WinoMT where gendered pronouns are replaced with the neutral form -- \textit{they/them} -- and any adjective/verb agreement is adjusted accordingly (see Figure~\ref{fig:neutralized_example}). This set enables a direct measurement of Prior Bias, \textit{i.e.}, a model’s default gender preference in the absence of explicit gender cues, thereby enriching the original framework.

Following prior work, we focus on English$\rightarrow$Italian, a contrastive pair where implicit source gender often must be made explicit in the target via agreement \citep{stahlberg}.

\section{Output-Level Analysis}\label{sec:Eval_MPA}
Our output-level analysis aims to answer the diagnostic question: \emph{When contextual gender cues are available, do models rely on them, or do they default to prior gender preferences?} We therefore combine three complementary metrics:
(i) Standard (Gender) Accuracy, from the integrated WinoMT pipeline,
(ii) Minimal Pair Accuracy (MPA) \citep{manna-etal-2025-paying}, and
(iii) \textbf{Prior Bias}, our novel measure.
Together, these allow us to attribute errors to either neglected cues, default gender assumptions, or both.

\subsection{Standard Gender Accuracy}\label{sec:winomt}
We first rely on the integrated WinoMT evaluation pipeline to measure how often the gender of the translated profession noun matches the gold label, as determined by the source pronoun. Input and target sentences are word-aligned using \texttt{fast\_align}\footnote{\href{https://github.com/clab/fast_align}{github.com/clab/fast\_align}}, and the gender of the aligned target term is extracted using \texttt{spaCy}'s\footnote{\href{https://spacy.io/}{spacy.io}} morphological analyzer. When no gender is detected, the sentence is assigned an \textsc{Unknown} label. Following the original pipeline, such cases are treated as errors, \textit{i.e.}, inaccurate gender realizations, and therefore penalize a model's accuracy score.

Table~\ref{tab:winomt_combined} reports standard (gender) accuracy metrics -- covering overall performance, as well as separate results for masculine and feminine referents -- across the regular, pro-, and anti-stereotypical subsets, for both our decoder-only models and the encoder-decoder systems previously studied in \citet{manna-etal-2025-paying}.

Two general tendencies hold across architectures: (i) models perform consistently better on masculine referents, and (ii) accuracy is higher in stereotypical contexts where the assigned gender aligns with societal expectations. However, what stands out is that decoder-only models do not generally outperform encoder-decoder systems on gender-related accuracy metrics, regardless of the larger size and architecture. 

Moreover, no improvement is observed between Llama2 and TowerBase, despite the latter having been reported to improve general translation quality \citep{alves2024tower}. In fact, TowerBase performs worse than the base model on anti-stereotypical cases. By contrast, TowerInstruct shows a substantial overall gain, which comes with feminine accuracy rising sharply from 23.3\% to 52.0\%. Yet these improvements are not uniform: while overall and feminine accuracies increase, performance on masculine referents decreases. This trade-off, absent from the previously studied encoder-decoder models \citep{manna-etal-2025-paying}, points to a redistribution rather than a uniform improvement.

\begin{table}[h!]
\centering
\small
\begin{tabular}{llccc}
\toprule
\textbf{Set} & \textbf{Model} & \textbf{Overall} & \textbf{Masc.} & \textbf{Fem.} \\
\midrule
\multicolumn{5}{c}{\textit{Decoder--Only Models}} \\
\midrule
REG   & Llama2        & 48.9\% & \textbf{80.8\%} & 23.3\% \\
      & TowerBase     & 47.4\% & 78.4\% & 22.6\% \\
      & TowerInstruct & \textbf{58.6\%} & 72.9\% & \textbf{52.0\%} \\
\midrule
PRO-S & Llama2        & 64.8\% & \textbf{90.2\%} & 39.5\% \\
      & TowerBase     & 65.2\% & 88.5\% & 41.9\% \\
      & TowerInstruct & \textbf{75.9\%} & 86.5\% & \textbf{65.4\%} \\
\midrule
ANTI-S & Llama2        & 40.1\% & \textbf{70.2\%} & 9.9\% \\
       & TowerBase     & 36.6\% & 64.7\% & 8.4\% \\
       & TowerInstruct & \textbf{49.7\%} & 58.4\% & \textbf{40.9\%} \\
\addlinespace
\midrule
\multicolumn{5}{c}{\textit{Encoder--Decoder Baselines (\citealt{manna-etal-2025-paying})}} \\
\midrule
REG   & OPUS-MT  & 42.6\% & 70.1\% & 20.6\% \\
      & NLLB-200 & 57.0\% & 79.6\% & 41.8\% \\
      & mBART    & \textbf{60.9\%} & \textbf{83.2\%} & \textbf{46.5\%} \\
\midrule
PRO-S & OPUS-MT  & 55.7\% & 77.3\% & 34.1\% \\
      & NLLB-200 & 74.9\% & 87.4\% & \textbf{62.5\%} \\
      & mBART    & \textbf{76.6\%} & \textbf{92.2\%} & 61.0\% \\
\midrule
ANTI-S & OPUS-MT  & 34.2\% & 59.1\% & 9.2\% \\
       & NLLB-200 & 47.3\% & 70.4\% & 24.2\% \\
       & mBART    & \textbf{54.0\%} & \textbf{71.9\%} & \textbf{35.9\%} \\
\bottomrule
\end{tabular}
\vspace{0.5em}
\caption{\small
Overall, masculine, and feminine (standard) accuracies on WinoMT for our decoder-only models and previously reported encoder-decoder systems \citep{manna-etal-2025-paying}, on the regular (REG), pro-stereotypical (PRO-S), and anti-stereotypical (ANTI-S) sets.}
\label{tab:winomt_combined}
\end{table}

\subsection{Minimal Pair Accuracy (MPA)}\label{sec:mpa}
Models can output the correct gender form without actually relying on the contextual cue, by defaulting to stereotypical associations. MPA~\citep{manna-etal-2025-paying} addresses this limitation by only "rewarding" context-sensitive predictions rather than simply correct ones. Minimal Pairs (see Fig.~\ref{fig:minimal_pair_example}) consist of two otherwise identical sentences that differ only in the gendered pronouns (\textit{he}/\textit{she}). A pair is considered accurate, only if the model produces the appropriately gendered translation for the profession \textit{in both cases}, therefore demonstrating it adapted based on the contextual cue (and thus did not simply rely on a stereotypical default).

\begin{figure}[h!]
\centering
\begin{LangBox}{Pro-S}
The analyst consulted with the \textbf{librarian} because \femalehl{\textbf{she}} knows a lot about books.
\end{LangBox}
\begin{LangBox}{Anti-S}
The analyst consulted with the \textbf{librarian} because \malehl{\textbf{he}} knows a lot about books.
\end{LangBox}
\vspace{0.75em}
\caption[]{\small{Minimal pair illustration. The pro-stereotypical case (top) assigns \textit{she} to \textit{librarian}, while the anti-stereotypical case (bottom) assigns \textit{he}. Accurate EN$\rightarrow$IT translations must adapt grammatical gender: \textit{la bibliotecaria} (\femalehl{f.}) vs.\ \textit{il bibliotecario} (\malehl{m.}).}}\label{fig:minimal_pair_example}
\end{figure}
\begin{table}[h!]
\centering
\small
\begin{tabular}{@{}lcccc@{}}
\toprule
\textbf{Model} & \textbf{MPA} & & \textbf{Pro-F} & \textbf{Pro-M} \\
\midrule
\multicolumn{5}{c}{\textit{Decoder--Only Models}} \\
\midrule
Llama2        & 13.1\% & & 73.4\% & 26.6\% \\
TowerBase     & 12.3\% & &  77.4\% & 22.6\% \\
TowerInstruct & \textbf{36.7\%} & & \textbf{50.4\%} & \textbf{49.6\%} \\
\addlinespace
\addlinespace
\midrule
\multicolumn{5}{c}{\textit{Encoder--Decoder Baselines (\citealt{manna-etal-2025-paying})}} \\
\midrule
OPUS-MT  & 6.1\% & & 82.3\% & 17.7\% \\
NLLB-200 & 30.2\% & & 69.1\% & 30.9\% \\
mBART    & \textbf{38.5\%} & & \textbf{61.9\%} & \textbf{38.1\%} \\
\bottomrule
\end{tabular}
\vspace{0.5em}
\caption{\small
Minimal Pair Accuracy (MPA) and its breakdown by gender and stereotype for decoder-only models and encoder-decoder baselines (results reported in \citet{manna-etal-2025-paying}). Pro-F and Pro-M indicate the proportion of accurate minimal pairs where the profession is stereotypically associated with women or men, respectively.}
\label{tab:mpa_combined}
\end{table}

As shown in Table~\ref{tab:mpa_combined}, MPA scores are considerably lower than standard Gender Accuracy (Table~\ref{tab:winomt_combined}), confirming that models often produce the correct gender form without relying on the contextual cue. Moreover, differences across decoder-only models are primarily driven by the instruction-tuned variant. Llama2 reaches a score of 13.1\%, which slightly decreases with TowerBase (12.3\%). By contrast, instruction tuning marks a clear shift, with TowerInstruct significantly outperforming the base model. When comparing our results to the previously studied encoder-decoder models \citep{manna-etal-2025-paying}, we observe that these achieve comparable or higher MPA values, indicating that state-of-the-art decoder-only LLMs are not inherently better at context-sensitive gender agreement despite their larger scale and broader pretraining.

We further break down our results by reporting the proportion of accurate minimal pairs in which the profession is stereotypically associated with the feminine (Pro-F) or masculine gender (Pro-M). Consistent with previous findings \citep{manna-etal-2025-paying}, both Llama2 and TowerBase exhibit a pronounced skew: most accurate pairs involve stereotypically feminine professions, whereas stereotypically masculine ones rarely appear in both gender forms (e.g., \textit{mechanic} is never rendered in its Italian feminine form, \textit{la meccanica}). TowerInstruct, however, diverges from this pattern by achieving an almost perfectly balanced split between Pro-F and Pro-M (50.4\% / 49.6\%). This pattern supports a shift away from a masculine default and toward more consistent cue integration.

\subsection{Prior Bias}\label{sec:prior}

To measure default gender preferences directly, we turn to our Neutral Set. We apply the same alignment and morphological analysis pipeline that supports WinoMT (Section~\ref{sec:winomt}) to extract the realized gender of the target profession. 

Since no explicit gender cue is present in this setting, there is no notion of correctness and the observed masculine/feminine distribution reflects the model’s \textbf{Prior Bias}. Accordingly, the metric is computed over instances in which a gender realization is detected, thereby excluding \textsc{Unknown} cases.

\begin{table}[h!]
\centering
\small
\begin{tabular}{lccc}
\toprule
\multicolumn{4}{c}{\textbf{Standard Accuracy}} \\
\midrule
\textbf{Model} & \textbf{Overall} & \textbf{Masculine} & \textbf{Feminine} \\
\midrule
Llama2        & 48.9\% & \textbf{80.8\%} & 23.3\% \\
TowerBase     & 47.4\% & 78.4\% & 22.6\% \\
TowerInstruct & \textbf{58.6\%} & 72.9\% & \textbf{52.0\%} \\
\addlinespace
\midrule
\multicolumn{4}{c}{\textbf{Diagnostic Evaluation}} \\
\midrule
 & \textbf{MPA} & \multicolumn{2}{c}{\textbf{Prior Bias}} \\
 \cmidrule(lr){3-4}
\textbf{Model} & & Masc. & Fem. \\
\midrule
Llama2        & 13.1\% & 85.3\% & 14.7\% \\
TowerBase     & 12.3\% & 85.6\% & 14.4\% \\
TowerInstruct & \textbf{36.7\%} & \textbf{75.2\%} & \textbf{24.8\%} \\
\bottomrule
\end{tabular}
\vspace{0.5em}
\caption{\small
Standard Accuracy reports traditional WinoMT metrics (Overall, Masculine, Feminine Accuracy). Diagnostic Evaluation combines (i) Minimal Pair Accuracy (MPA); and (ii) \textbf{Prior Bias}, capturing the proportion of masculine vs.\ feminine gendered forms on the neutralized set, \textit{i.e.}, in the absence of explicit cues.}
\label{tab:accuracy_vs_diagnostic}
\end{table}

Table~\ref{tab:accuracy_vs_diagnostic} summarizes our results across all evaluated metrics, combining standard WinoMT accuracies with both existing and extended diagnostic measures. 
While standard Gender Accuracy and MPA already pointed to an overall masculine default, the neutral-set evaluation makes this explicit: both Llama2 and TowerBase overwhelmingly default to the masculine gender ($\sim$85\%), showing no evidence of a preference shift. By contrast, in line with its higher MPA and more balanced Pro-F/Pro-M distribution, TowerInstruct reduces this default and generates feminine forms more frequently ($\sim$25\%). Although a masculine prior remains dominant overall, instruction tuning is here associated with both stronger cue integration and a marked reduction in masculine Prior Bias.

\section{Attention-Based Analysis}
While output-level metrics reveal whether models correctly resolve gender, they do not show how cues are processed internally. 
In this section, we examine the models' internals by analyzing the pattern of self-attention -- the core mechanism of context mixing in Transformers \citep{mohebbi-etal-2023-quantifying} -- to understand how contextual cues are integrated into the representation of the target word during the translation process.

\subsection{Setup} 

For each layer and head, we extract the self-attention weight allocated to the gender cue at the time step in which the target word is generated. 
Since target words are often split into multiple tokens, we implement a matching strategy to retrieve the full span of subwords for each tokenized target, explicitly including inflected articles (\textit{il, lo, la}, etc.), which are the first locus of gender realization in Italian. This step is language-specific but easily adaptable to other target languages by either modifying or omitting the gendered article component. 

Attention scores are then averaged across all subwords, and aggregated over accurate minimal pairs, \textit{i.e.}, cases where the model correctly adapted the target to the pronouns (see Section~\ref{sec:mpa}). Because these cues are the only explicit gender indicators in the source, restricting the analysis to accurate pairs allows us to identify where -- across layers and heads -- gender information is encoded in the representation of the target words. To ensure comparability across such aggregated scores, we further restrict the analysis to the minimum number of accurate minimal pairs observed across systems ($n=195$).

\begin{figure*}[h!]
\centering

\begin{subfigure}[t]{0.9\textwidth}
\centering
\includegraphics[
  width=\textwidth,
  keepaspectratio
]{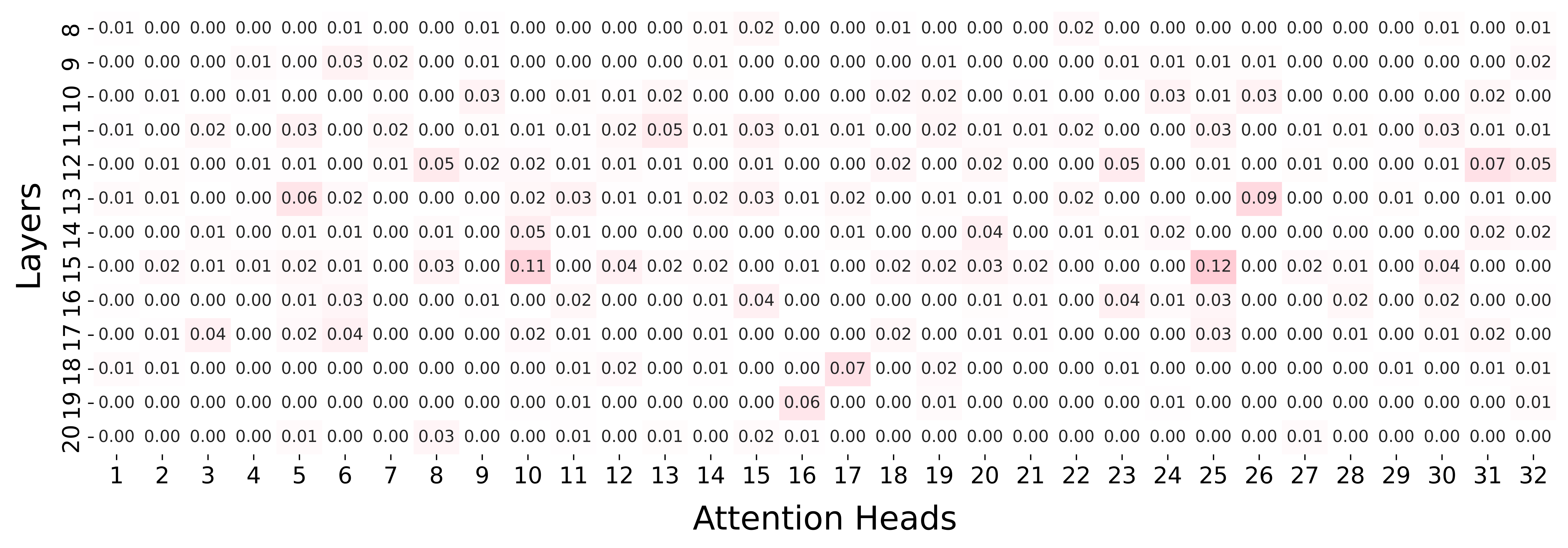}
\caption{Llama2}
\label{fig:att_llama2_fem}
\end{subfigure}

\vspace{0.25em}

\begin{subfigure}[t]{0.9\textwidth}
\centering
\includegraphics[
  width=\textwidth,
  keepaspectratio,
]{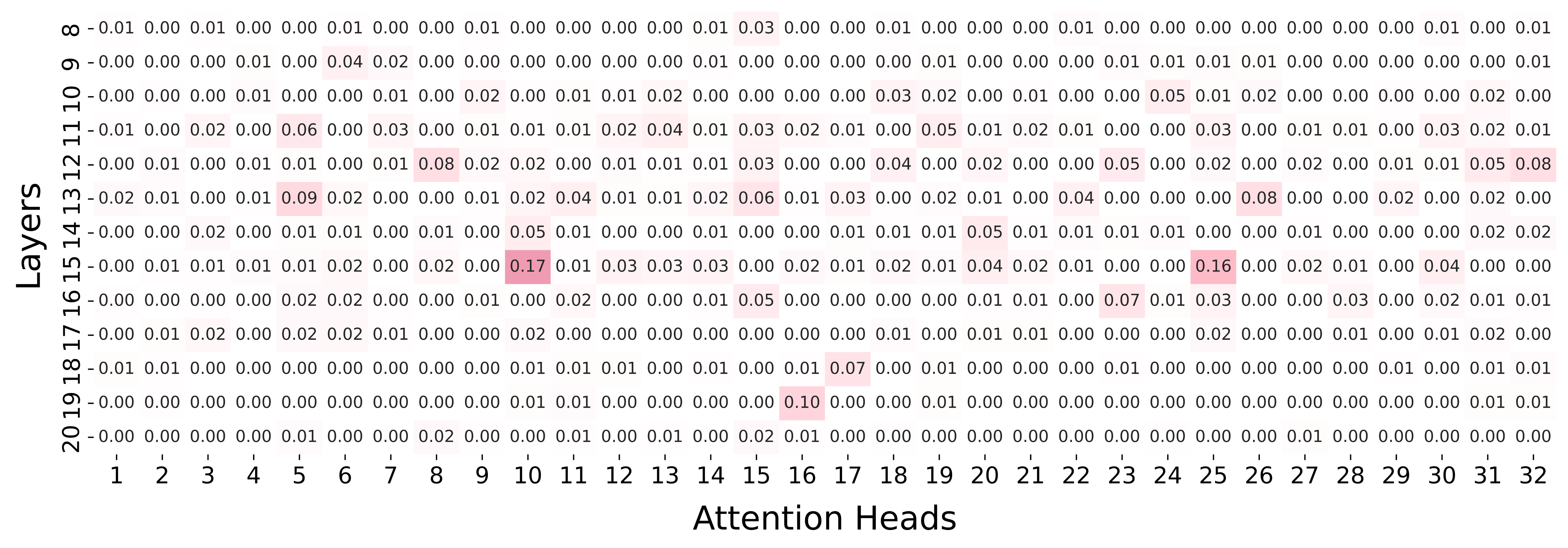}
\caption{TowerBase}
\label{fig:att_towerbase_fem}
\end{subfigure}

\vspace{0.25em}

\begin{subfigure}[t]{0.9\textwidth}
\centering
\includegraphics[
  width=\textwidth,
  keepaspectratio,
]{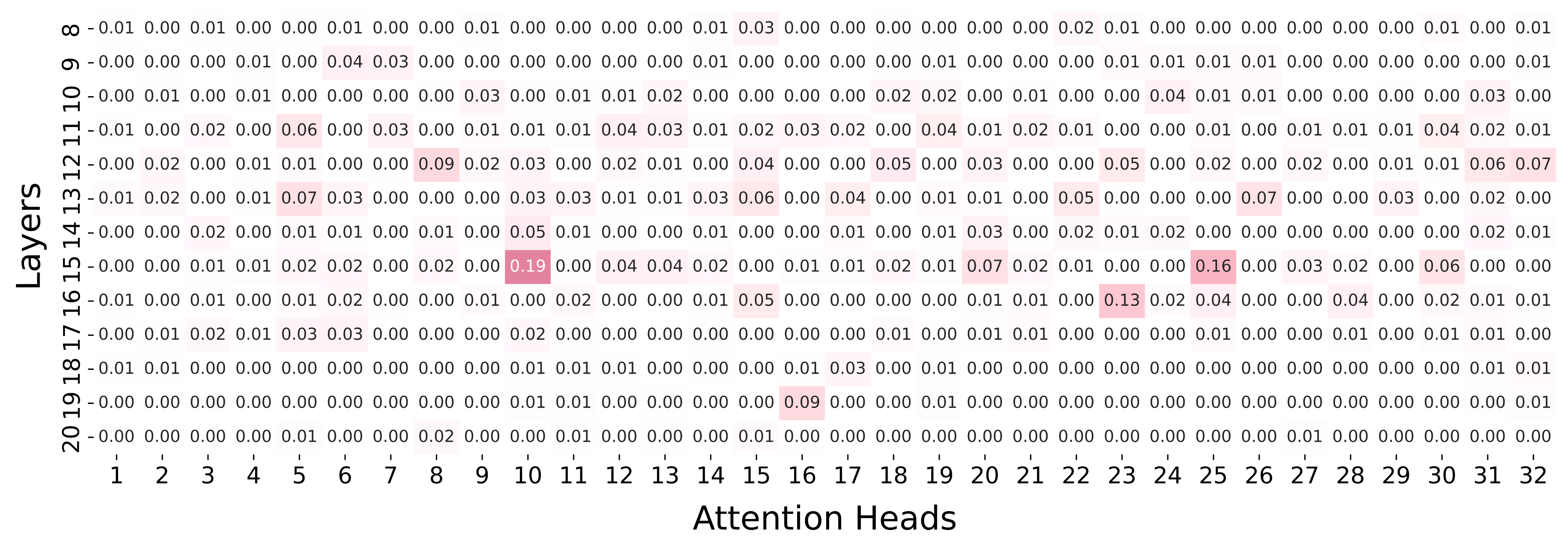}
\caption{TowerInstruct}
\label{fig:att_towerinstruct_fem}
\end{subfigure}
\caption{\small
Attention from the translated profession noun to the \femalehl{feminine cue} across layers (y-axis) and heads (x-axis). Values represent the average attention weight assigned to the cue across the minimum number of accurate minimal pairs observed across models ($n=195$). We focus on intermediate layers (8--20), where attention is strongest; earlier and later layers exhibit near-zero values. Same color scale applied; darker shades $\rightarrow$ higher average attention.}
\label{fig:att_fem_all_vertical}
\end{figure*}

After a preliminary inspection, we observe that attention to masculine pronouns is generally weaker, whereas feminine pronouns trigger stronger attention patterns. This suggests that our decoder-only models, like the encoder-decoder ones analyzed in \citet{manna-etal-2025-paying}, must actively integrate contextual information to override a masculine default -- aligning with the strong masculine Prior Bias identified in Section~\ref{sec:prior}. We therefore focus on \textbf{feminine cues} in \textbf{accurate minimal pairs}.

\subsection{Attention Patterns}\label{att_analysis}
We observe that certain attention heads consistently attend to gender cues, with feminine cues often eliciting stronger activations, likely compensating for their lower frequency in training data. This is consistent with previous findings \citep{manna-etal-2025-paying}, where we similarly reported weaker and more diffuse attention to masculine cues, and stronger, more localized attention to feminine ones for encoder-decoder models.

Figure~\ref{fig:att_fem_all_vertical} displays attention heatmaps for feminine cues across layers (y-axis) and heads (x-axis).\footnote{We focus on intermediate layers (8--20), where attention to the pronoun is strongest.}

Specific heads become increasingly responsive when the base model (Llama2) undergoes post-training. For instance, in Layer 15, attention in Heads 10 and 25 rises steadily from $0.11$ and $0.12$ to $0.17$ and $0.16$ in TowerBase, with Head 10 further increasing to $0.19$ in TowerInstruct. 

Our analysis offers deeper insights into not only where but also when model internals shape gender disambiguation performance in machine translation.
The most pronounced increase in self-attention weights can be observed between Llama2 and TowerBase, whereas the most marked reduction of the inherent masculine Prior Bias only emerges with TowerInstruct (cf. Section~\ref{sec:prior}).
This suggests that, although continued pretraining on MT-related data enables the base model to better identify and incorporate contextual gender cues into the representation of the target words, these improvements are not directly reflected in the model's output. In fact, it is only post instruction tuning that the model appears to effectively leverage these learned attention signals to mitigate its masculine Prior Bias.
This aligns with \citet{Choshen2019OnTW}, who observe that post-training practices tend to amplify latent signals already learned by the model, rather than introducing entirely new ones.



\subsubsection{Sanity Checks}
While attention represents a popular choice to interpret and visualize token interactions across layers and heads in MT research, its reliability as explanatory signal has been widely debated \citep{jain-wallace-2019-attention, bibal-etal-2022-attention}. We therefore complement our analysis with targeted sanity checks.

\paragraph{Prompt Attention} 
Although we keep the translation instruction identical for all models (Section~\ref{sec:prompt}), the TowerInstruct model includes additional special tokens in its chat template. This raises the question of whether attention to gender cues can be directly compared across models.
To check this, we measure the total attention each model assigns to its prompt tokens (see Figure~\ref{fig:prompt_templates}), \textit{i.e.}, the sum of the attention weights associated to all prompt tokens when generating the target word.
We see a highly consistent value across models ($\sim$$0.77$--$0.78\%$), indicating that our direct comparison is fair as the main context receives equivalent total attention across models.

\begin{table}[h!]
    \centering
    \small
    \begin{tabular}{lcccc}
    \toprule
         &  \multicolumn{2}{c}{\textbf{Target (L15)}} & \multicolumn{2}{c}{\textbf{Secondary (L15)}} \\
         \cmidrule(lr){2-3} \cmidrule(lr){4-5}
         \textbf{Model} & \textbf{H10} & \textbf{H25} & \textbf{H10} & \textbf{H25} \\
         \midrule
         Llama2 & 0.11 & 0.12 & 0.01 & 0.02 \\
         TowerBase & 0.17 & 0.16 & 0.02 & 0.01 \\
         TowerInstruct & 0.19 & 0.16 & 0.04 & 0.03 \\
         \bottomrule
    \end{tabular}
    \vspace{0.5em}
    \caption{\small Attention to the \femalehl{feminine cue} in Layer 15, Heads 10 and 25 when generating the translation of our target v.s.\ our secondary (non-coreferent) entity.}
    \label{tab:l15_comp}
\end{table}

\paragraph{Secondary Entity} 
We further test whether models might attend to pronouns generically, rather than to resolve the gender of the coreferent entity. Since each WinoMT sentence includes a secondary (non-coreferent) entity (e.g., \textit{The doctor asked the \textbf{nurse} to help her.}), we annotate their positions and measure the attention the model allocates to the cue when generating their corresponding translations (see Appendix~A for the corresponding heatmaps).
We observe that this is consistently low ($\leq$$0.04$), suggesting that models attend to pronouns selectively, only when these are relevant for gender  disambiguation. Table~\ref{tab:l15_comp} reports these values for our prominent heads, Heads 10 and 25 at Layer~15, and compares them with the attention from our target in Figure~\ref{fig:att_fem_all_vertical}.

\section{Conclusion}
We extend the original evaluation framework \citep{manna-etal-2025-paying} by (i) enriching it with an explicit measure of \textbf{Prior Bias}, \textit{i.e.}, a model's default gender preference, and (ii) applying it to \textbf{decoder-only models} in both their general-purpose and translation-tuned variants.

We find that decoder-only models, which underpin the current state-of-the-art, do not consistently outperform encoder-decoder systems \citep{manna-etal-2025-paying}. This observation resonates with recent evidence that decoder-only LLMs tend to underperform encoder- or encoder-decoder models on tasks requiring more contextual disambiguation (e.g., word meaning tasks~\citep{qorib2024decoder}). Nonetheless, similar overall tendencies persist: all models perform better on masculine referents and in stereotypical contexts. Moreover, MPA consistently falls below standard Gender Accuracy, revealing that even when models produce the expected target gender, they do not consistently rely on contextual cues. While all evidence points to a default-to-masculine reasoning process -- which aligns with prior work \citep{jumelet-etal-2019-analysing, birth_of_bias} -- this tendency is now explicitly captured and quantified through our Prior Bias metric. Together, these results help contextualize systematic gender errors in MT, tracing them to ignored contextual cues and pronounced gender priors. 

At the same time, our setup enables a controlled comparison across TowerLLM's post-training stages. While continued pretraining offers little benefit for gender disambiguation, instruction tuning is associated with stronger cue integration - reflected in higher MPA, improved feminine accuracy, and a marked reduction in masculine Prior Bias. Ultimately, our framework provides a diagnostic tool to better understand how gender is encoded and resolved in translation models, and to inform interventions that can improve model behavior. Where traditional gender bias mitigation approaches remain difficult to scale \citep{savoldi2025decade}, instruction tuning offers a promising direction towards more contextually aware and gender-fair translation behavior, especially considering the persistent challenge of ensuring unbiased, high-quality training data in large-scale MT and LLMs.

\section{Limitations}
While our work strengthens the WinoMT framework \citep{winomt}, it also inherits some of its limitations. First, the analysis remains constrained to a binary gender distinction, thereby excluding non-binary and gender-neutral alternatives, which are of social and linguistic relevance but remain underrepresented in most current NLP resources. Developing datasets and extending evaluation frameworks beyond binary gender thus remains an important direction for future work. 

Second, the evaluation pipeline relies on two automatic processing steps---\texttt{fast\_align} and \texttt{spaCy}---which inevitably introduce some degree of noise into the analysis. In addition to erroneous gender assignments, there are instances in which no gender can be (automatically) detected, resulting in an \textsc{Unknown} label. These may arise from incorrect translations (e.g., incoherent output or omission of the target entity), alignment errors, or inherent limitations of the morphological analyzer. Nonetheless, such cases are rarely explicitly reported, as the pipeline is often applied without further inspection. 
\begin{table}[h!]
\centering
\small
\begin{tabular}{lc}
\toprule
\textbf{Model} & \textbf{Unknown (\%)} \\
\midrule
Llama2        & 9.4\% \\
TowerBase     & 10.3\% \\
TowerInstruct & 12.8\% \\
\bottomrule
\end{tabular}
\vspace{0.5em}
\caption{\small
Percentage of \textsc{Unknown} gender assignments for each model on the regular WinoMT set, \textit{i.e.}, instances where the automatic pipeline failed to detect a gender.}
\label{tab:unknown_rates}
\end{table}

To quantify this phenomenon, we report the proportion of \textsc{Unknown} cases in our analysis (Table~\ref{tab:unknown_rates}). Although these are broadly comparable across models and do not alter our qualitative conclusions, even under the conservative assumption that all \textsc{Unknown} cases correspond to correct gender realizations, their frequency deserves attention and may reflect both model-specific behavior and language-specific ambiguity. 
While this is beyond the scope of this work, a more fine-grained analysis to disentangle genuine gender errors from pipeline-specific or cross-linguistic artifacts is left for future investigation.

Beyond dataset- and evaluation-related considerations, we acknowledge that the attention-based analysis provides correlational rather than causal evidence. Complementing it with causal interpretability techniques would allow for a more direct assessment of the role of gender cues in translation decisions. Finally, while instruction tuning emerges as a promising avenue for gender bias mitigation in MT, it remains unclear which specific aspects of the process drive the observed improvements. Further investigation is needed to understand how post-training practices can contribute to the development of more transparent bias mitigation strategies.


\section{Bias Statement}
We understand gender bias in MT as a model's systematic tendency to rely on learned statistical regularities rather than consistently relying on contextual information for gender disambiguation, as previously defined \citep{manna-etal-2025-paying}. 
Our analysis is restricted to non-ambiguous contexts, in which gender is explicitly signaled in the source -- typically through gendered pronouns referring to human entities -- representing one specific subtype of gender bias. In these cases, translations often appear to be driven by prior gender associations rather than contextual evidence. This can appear either as stereotypical bias, where models tend to only produce feminine translations when the occupation is socially associated with women (e.g., \textit{housekeeper} becomes \textit{la governante}), or as a masculine default, where models revert to masculine forms regardless of the cue, reinforcing a male-as-norm pattern \citep{danesi-2014}. Such behaviors can lead to both representational harm, by perpetuating traditional gender roles, and allocational harm, by systematically underrepresenting women in male-dominated professions \citep{blodgett-etal-2020-language}.

Because the WinoMT dataset \citep{winomt} encodes gender as a binary variable, and its underlying linguistic tools follow the same assumption, our analysis is restricted to masculine and feminine gender realizations. We acknowledge that this is a major limitation, as gender exists along a spectrum rather than a binary divide. However, there is still no broadly adopted methodology to evaluate non-binary gender bias in MT. Developing evaluation frameworks that better reflect gender diversity and inclusivity remains an important direction for future research.

\section{Bibliographical References}\label{sec:reference}
\bibliographystyle{lrec2026-natbib}
\bibliography{lrec2026-example}

\clearpage
\onecolumn
\appendix
\section*{Appendix A. Attention to Secondary Entities}

\begin{figure*}[h!]
\centering

\begin{subfigure}[t]{0.9\textwidth}
\centering
\includegraphics[
  width=\textwidth,
  keepaspectratio,
]{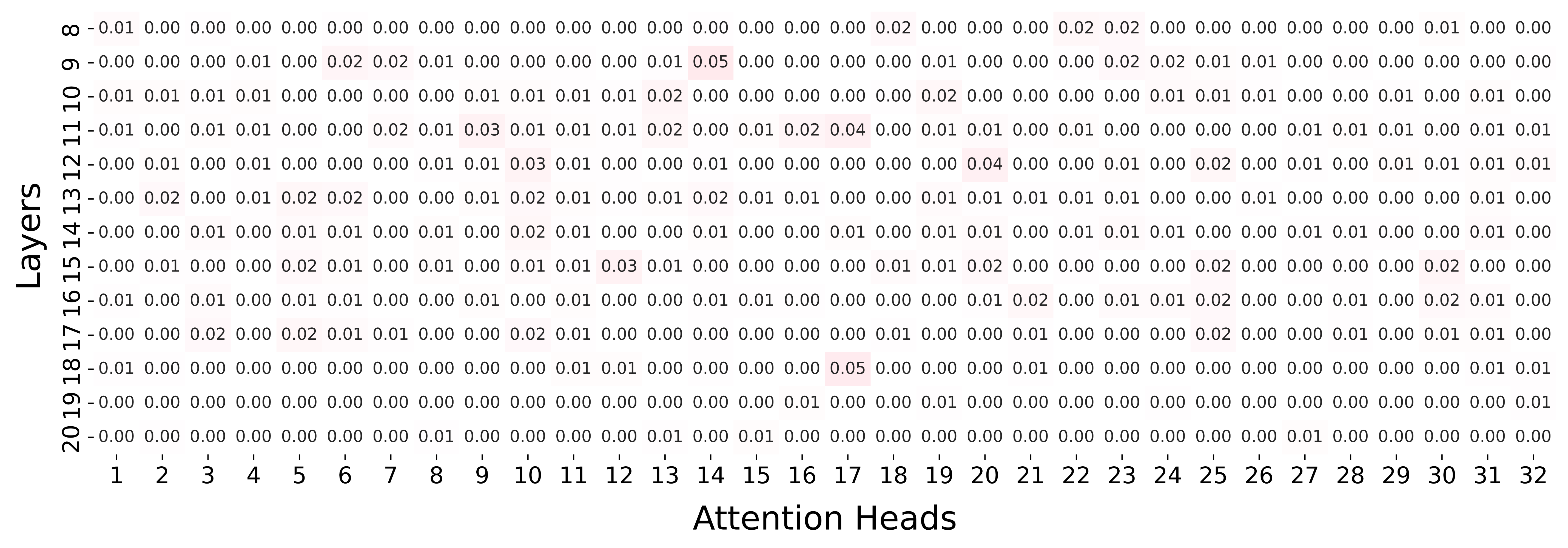}
\caption{Llama2}
\end{subfigure}

\vspace{0.25em}

\begin{subfigure}[t]{0.9\textwidth}
\centering
\includegraphics[
  width=\textwidth,
  keepaspectratio,
]{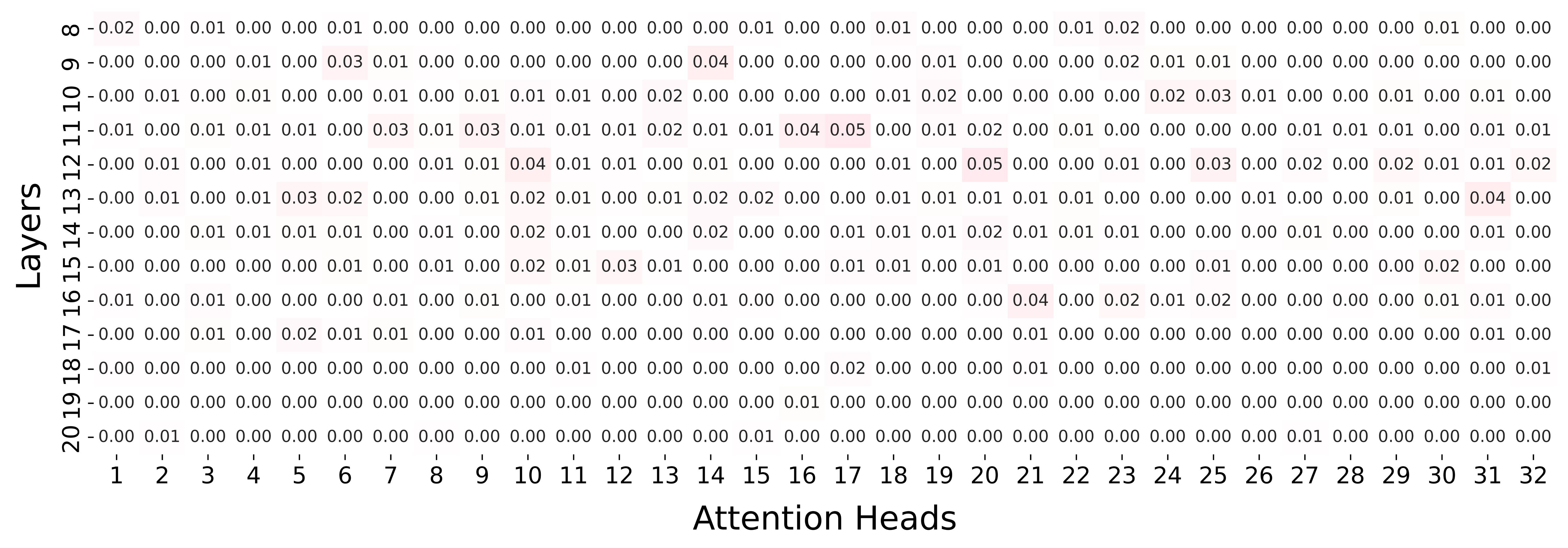}
\caption{TowerBase}
\end{subfigure}

\vspace{0.25em}

\begin{subfigure}[t]{0.9\textwidth}
\centering
\includegraphics[
  width=\textwidth,
  keepaspectratio,
]{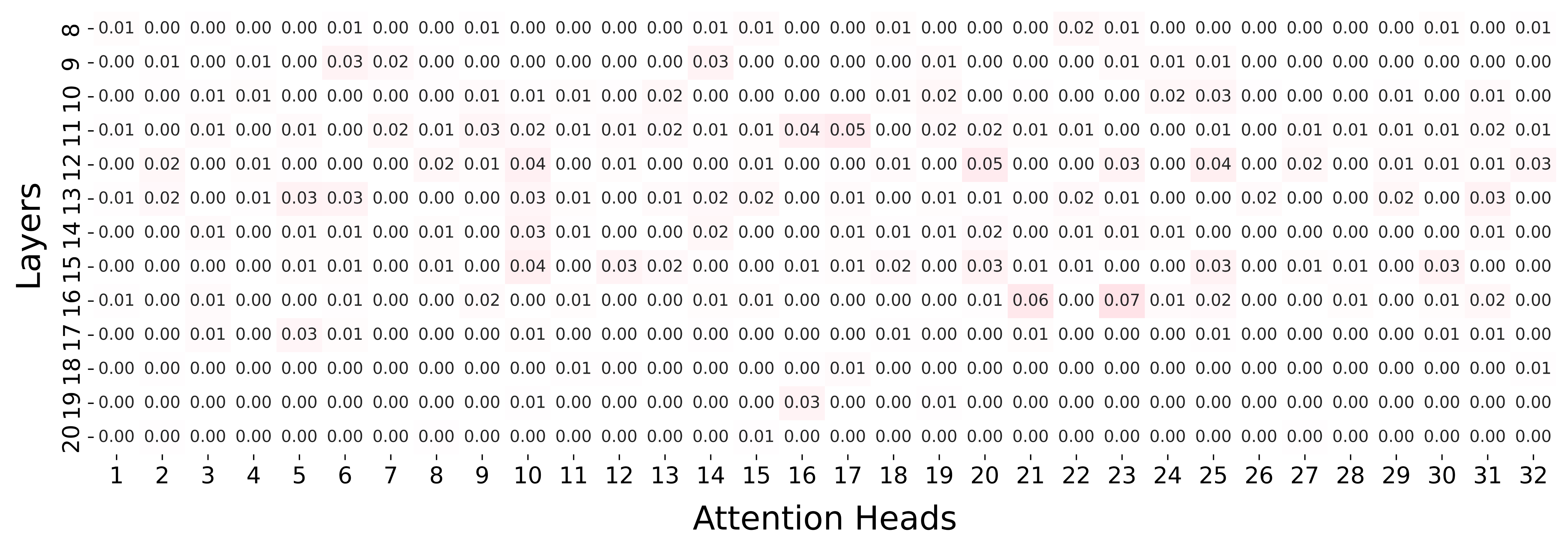}
\caption{TowerInstruct}
\end{subfigure}

\caption{\small
Attention from the translated secondary (non-coreferent) entity to the \femalehl{feminine cue} across layers (y-axis) and heads (x-axis).  Values represent the average attention weight assigned to the cue across accurate minimal pairs ($n=195$). Same color scale applied; darker shades $\rightarrow$ higher average attention.}
\label{fig:att_secondary_all_vertical}
\end{figure*}


\end{document}